\documentclass[conference]{IEEEtran}
\IEEEoverridecommandlockouts
\usepackage{cite}
\usepackage{amsmath,amssymb,amsfonts}
\usepackage{algorithmic}
\usepackage{graphicx}
\usepackage{textcomp}
\usepackage{xcolor}
\usepackage{url}

\usepackage{subfig}
\usepackage{color}
\usepackage{tikz}

\tikzset{layer/.style={fill=blue!10,auto,align=center,thick,draw,minimum width=1.1cm,minimum height=.5cm}}

\def\BibTeX{{\rm B\kern-.05em{\sc i\kern-.025em b}\kern-.08em
    T\kern-.1667em\lower.7ex\hbox{E}\kern-.125emX}}
\begin{document}

\title{Learning Image Relations with \\ Contrast Association Networks}

\author{\IEEEauthorblockN{Yao Lu}
\IEEEauthorblockA{\textit{ANU, Data61, ACRV}\\
yaolubrain@gmail.com}
\and
\IEEEauthorblockN{Zhirong Yang}
\IEEEauthorblockA{
\textit{NTNU}\\
zhirong.yang@ntnu.no}
\and
\IEEEauthorblockN{Juho Kannala}
\IEEEauthorblockA{
\textit{Aalto University}\\
juho.kannala@aalto.fi}
\and
\IEEEauthorblockN{Samuel Kaski}
\IEEEauthorblockA{
\textit{Aalto University}\\
samuel.kaski@aalto.fi}
}

\maketitle

\begin{abstract}
Inferring the relations between two images is an important class of tasks in computer vision. Examples of such tasks include computing optical flow and stereo disparity. We treat the relation inference tasks as a machine learning problem and tackle it with neural networks. A key to the problem is learning a representation of relations. We propose a new neural network module, contrast association unit (CAU), which explicitly models the relations between two sets of input variables. Due to the non-negativity of the weights in CAU, we adopt a multiplicative update algorithm for learning these weights. Experiments show that neural networks with CAUs are more effective in learning five fundamental image transformations than conventional neural networks.
\end{abstract}

\section{Introduction}

Neural networks, especially convolutional neural network (CNN) \cite{lecun1998gradient}, have been successfully applied in many computer vision tasks such as
object recognition \cite{ciresan2012multi,krizhevsky2012imagenet}. A key to this success is that the neural networks allow appearance representation which is invariant of several image transformations such as small translation.

Another important class of tasks in computer vision is the inference of relations between two images. Two images can be related by object motion, camera motion or environmental factors such as lighting change. For these problems, instead of aiming for invariance of appearance changes, we want to detect and estimate appearance changes. For example, given two consecutive video frames, we want to infer the movement of each pixel from one image to the other (optical flow). For another example, given two images taken by a camera on a moving robot, we want to infer the ego-motion of the robot (visual odometry).

The traditional approach to perform the relation inference tasks is knowledge-based. 
By acquiring knowledge of a task and making reasonable assumptions for simplification, one designs an algorithm to perform the inference. A classic example is the Horn-Schunck algorithm for computing optical flow \cite{horn1981determining}. However, in situations where we lack sufficient knowledge or the assumptions fail, the knowledge-based approach may not perform well. For example, the Horn-Schunck algorithms assumes brightness constancy of the moving pixels. This assumption can be violated by many factors such as occlusion, shading and noise.

An alternative approach to solve the relation inference problem is learning-based \cite{memisevic2013learning}. From a collection of training data, we aim to learn a function
\begin{align}
\mathbf{z} = F(\mathbf{x},\mathbf{y})
\end{align}
such that given two images $\mathbf{x}$ and $\mathbf{y}$ as inputs, the function will output their relation variables $\mathbf{z}$. The relation variables can be rotation angle, motion field, affine transformation parameters, etc. Compared to the knowledge-based approach, the learning-based approach does not require sufficient knowledge or assumptions but a large amount of training images with ground-truth relation variables. 
When it is difficult to obtain the ground-truth for real world images, one can often resort to synthetic images rendered by graphics engines. An example is the Sintel dataset for learning optical flow  \cite{butler2012naturalistic}.
Recently, the learning-based approach has been adopted to compute optical flow \cite{fischer2015flownet,thewlis2016fully,ranjan2016optical,ilg2016flownet},
 stereo disparity \cite{luo2016efficient}, camera motion \cite{ummenhofer2016demon} and visual odometry \cite{konda2015learning}. 
Some of the results are competitive to the knowledge-based methods. 
Note that relation learning can also serve as supervision for learning appearance representation, as demonstrated in learning ego-motion \cite{agrawal2015learning,jayaraman2015learning} and robot actions \cite{pinto2016curious}.

Additionally, there are methods combining knowledge-based and learning-based approaches  \cite{zagoruyko2015learning,zbontar2016stereo,shaked2016improved,bailer2016cnn}.
In these methods, a neural network is trained to match two image patches and a knowledge-based post-processing is applied on the matching results to output the relation variables. In this paper, we focus on the pure learning-based approach, that is, neural networks are trained in an end-to-end manner, given raw images as inputs and ground-truth relation variables as targets.
Although a relation learning model can also be trained unsupervisedly \cite{memisevic2010learning,konda2013unsupervised,yu2016back}, we restrict our discussion mainly to supervised learning. 

While both object recognition and relation inference can be treated as supervised learning tasks, the two tasks have much difference in nature. 
Object recognition aims for invariance of several image transformations (e.g. translation, rotation and scaling) but relation inference aims for equivariance of these transformations. For example, the conventional CNN with a pooling operation is known to be invariant of small translation. This property is suitable for object recognition but not for motion detection, whose goal is to estimate the translation. This difference should be kept in mind when designing a relation learning model.

In this paper, we propose a new relation unit, \textit{contrast association unit} (CAU). We show that CAUs are suitable for relation learning tasks with analysis and experiments. We adopt a multiplicative update algorithm for learning the non-negative weights in CAUs. The multiplicative update algorithm is compatible with gradient descent algorithms for unconstrained weights. The whole neural network can be trained in an end-to-end manner.

Next, in Section \ref{sec:arch}, we outline the general neural network architecture for relation learning. In Section \ref{sec:cau}, we introduce our proposed relation unit CAU. In Section \ref{sec:learning}, we present the multiplicative update algorithm for learning the non-negative weights in CAUs. In Section \ref{sec:related_work}, we discuss models related to CAU. In Section \ref{sec:experiments}, we present the experiments on the five relation learning tasks.

\section{Architecture}\label{sec:arch}

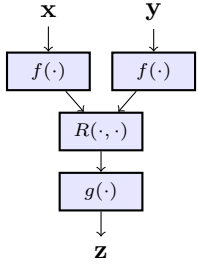
\begin{figure}[t!]
\centering
\begin{tikzpicture}
\node (x) at (-0.7,0) {$\mathbf{x}$}; 
\node (y) at (0.7,0)  {$\mathbf{y}$}; 
\node (z) at (0,-3.2)  {$\mathbf{z}$}; 
\node [layer](f1) at (-0.7,-0.8) {\scriptsize{$f(\cdot)$}}; 
\node [layer](f2) at (0.7,-0.8) {\scriptsize{$f(\cdot)$}}; 
\node [layer](cat) at (0,-1.6) {\scriptsize{$R(\cdot,\cdot)$}}; 
\node [layer](g) at (0,-2.4) {\scriptsize{$g(\cdot)$}}; 
\path[->] (x) edge node {} (f1);	
\path[->] (y) edge node {} (f2);	
\path[->] (f1) edge node {} (cat);
\path[->] (f2) edge node {} (cat);
\path[->] (cat) edge node {} (g);
\path[->] (g) edge node {} (z);
\end{tikzpicture}
\caption{Neural network architecture for relation learning}
\label{architecture}
\end{figure}

As illustrated in Figure \ref{architecture}, the general neural network architecture of many relation learning models \cite{memisevic2010learning,fischer2015flownet,agrawal2015learning} can be described as 
\begin{alignat}{2}
\mathbf{a} &= f(\mathbf{x}), \quad  \mathbf{h} &&= R(\mathbf{a},\mathbf{b}), \nonumber\\
\mathbf{b} &= f(\mathbf{y}), \quad  \mathbf{z} &&= g(\mathbf{h}),
\end{alignat}
where $\mathbf{x}$ and $\mathbf{y}$ are the inputs (images), $\mathbf{z}$ are the targets (relation variables), $f(\cdot)$ is the feature extraction units, 
$R(\cdot,\cdot)$ is the relation units and $g(\cdot)$ is the decoding units. $f(\cdot)$ and $g(\cdot)$ can be parametric functions such as neural networks. When the relation of two images is on the pixel-level (e.g. affine transformation), $f(\cdot)$ can be the identity function such that the relation units can directly apply on the image pixels. When the relation units $R(\cdot,\cdot)$ can directly output relation variables instead of a hidden representation, $g(\cdot)$ can also be the identity function.
For example, in \cite{memisevic2010learning}, both $f(\cdot)$ and $g(\cdot)$ are the identity function.
For another example, in the simple version of FlowNet \cite{fischer2015flownet}, $f(\cdot)$ is the identity function and $g(\cdot)$ is a CNN while in the complex version, both $f(\cdot)$ and $g(\cdot)$ are CNNs. 
The main focus of this paper is the relation units $R(\cdot,\cdot)$. 
A suitable relational representation can reduce the sample complexity and model complexity of learning $g(\cdot)$. We present two common types of relation units in below.

\subsection{Concatenation Units}
Concatenation units are defined as 
\begin{align}
\mathbf{h} =  [\mathbf{a}\ \ \mathbf{b}].
\end{align}
For this simple representation, $g(\cdot)$ is solely responsible for learning the relations between $\mathbf{a}$ and $\mathbf{b}$.
Concatenation units have been used in learning optical flow \cite{fischer2015flownet} and ego-motion \cite{agrawal2015learning}.

\subsection{Bilinear Units}
Bilinear units have been previously proposed and developed  \cite{hinton1981parallel,olshausen1993neurobiological,tenenbaum2000separating}.
They are defined as, for the $k$-th unit, 
\begin{align}
h_k = \sum_{ij}W_{ijk}a_ib_j = \mathbf{a}^T \mathbf{W}_k \mathbf{b},
\label{bilinear}
\end{align}
where $\mathbf{W}_k$ is the parameters to be learned.
A bilinear unit models the pair-wise multiplicative intersection between two sets of input variables. It is equivalent to inner product when $\mathbf{W}_k$ is the identity matrix and equivalent to outer product since (\ref{bilinear}) can be written as $h_k = \text{vec}(\mathbf{W}_k)^T\text{vec}(\mathbf{ab}^T)$, where $\text{vec}(\cdot)$ denotes vectorization. Bilinear units have been used in learning image transformations \cite{memisevic2010learning,rocco2017convolutional}.

\section{Contrast Association Units} \label{sec:cau}

We propose a new relation unit, \textit{contrast association unit} (CAU), which associates two sets of input variables. CAUs are defined as, for the $k$-th unit,
\begin{align}
h_k = \frac{1}{2}\sum_{ij}W_{ijk}(a_i-b_j)^2,
\label{cau}
\end{align}
where $W_{ijk} \geq 0$. 
Each CAU can be interpreted as a weighted sum of mismatches between $\mathbf{a}$ and $\mathbf{b}$. The non-negative constraint on the weight matrices is indispensable since the mismatches should be non-negative to be accumulated. Otherwise, positive mismatches and negative mismatches would cancel each other.

Compared to concatenation and bilinear units, CAUs have two advantages: (1) as the name suggests, CAUs model the contrast between two sets of variables such that 
$R(\mathbf{a}+c,\mathbf{b}+c) = R(\mathbf{a},\mathbf{b})$,
where $c$ is a scalar applied on each element of a vector. This property is desirable for relation inference. For example, when $\mathbf{a}$ and $\mathbf{b}$ are the raw pixels, their relations should not be affected by their absolute pixel intensity level. (2) CAUs 
represent relations more explicitly. Each CAU stands for the matching error of a certain relation encoded by the weight matrix. An example is given in Section \ref{example}. This interpretablity also inspires the use of competition among CAUs as described below.

\subsection{Competition}
We apply a competition mechanism among CAUs. The competition encourages the unit standing for the minimum matching error to pop out. As a result, the relation representation can be more easily read-out by the decoding units. 
A classic competition mechanism is the winner-take-all (WTA), defined as
\begin{align}
h_k' = 
\begin{cases}
1,\quad \text{if } h_k =\min(\mathbf{h}),   \\
0,\quad \text{otherwise.}
\end{cases}
\end{align}
WTA is of conceptual interest, which is demonstrated in Section \ref{example}. However, WTA is not differentiable and discards too much information.
In practice, we can use softmin competition, defined as 
\begin{align}
{h}'_k = \frac{e^{-h_k}}{\sum_ie^{-h_i}}.
\end{align}
In all our experiments, adding the softmin competition significantly improves the results of neural networks with CAUs.

\subsection{Example}\label{example}
To understand how neural networks with CAUs represent relations, let us consider a simple example of translation detection.
Let
\begin{align}
\mathbf{a} &= (c_1,c_2,c_3,c_4,c_5),  \\
\mathbf{b}_1 &= (c_2,c_3,c_4,c_5,c_6), \\
\mathbf{b}_2 &= (c_1,c_2,c_3,c_4,c_5), \\
\mathbf{b}_3 &= (c_0,c_1,c_2,c_3,c_4), 
\end{align}
where $\{c_i\}$ are arbitrary numbers. Let $z \in \{-1,0,1\}$ be the translation variable and 
\begin{align}
\mathbf{b} = 
\begin{cases}
\mathbf{b}_1, & \text{if } z=-1, \\
\mathbf{b}_2, & \text{if } z=0, \\
\mathbf{b}_3, & \text{if } z=1. 
\end{cases}
\end{align}
Denote by $\mathbf{D}(\mathbf{a},\mathbf{b})$ the matrix of pair-wise squared differences of the elements in  $\mathbf{a}$ and $\mathbf{b}$. The element of index $(i,j)$ in $\mathbf{D}(\mathbf{a},\mathbf{b})$ is $(a_i-b_j)^2$. Then we have
\begin{align}
\underbrace{
  \scriptsize{
  \begin{bmatrix}
   * & * & * & * & * \\
   0 & * & * & * & * \\
   * & 0 & * & * & * \\
   * & * & 0 & * & * \\        
   * & * & * & 0 & * \\     
  \end{bmatrix}}
}_{\mathbf{D}(\mathbf{a},\mathbf{b}_1)},
\underbrace{
  \scriptsize{
  \begin{bmatrix}
   0 & * & * & * & * \\
   * & 0 & * & * & * \\
   * & * & 0 & * & * \\        
   * & * & * & 0 & * \\    
   * & * & * & * & 0        
  \end{bmatrix}}
}_{\mathbf{D}(\mathbf{a},\mathbf{b}_2)},
\underbrace{
  \scriptsize{
  \begin{bmatrix}  
   * & 0 & * & * & * \\
   * & * & 0 & * & * \\        
   * & * & * & 0 & * \\    
   * & * & * & * & 0 \\
   * & * & * & * & *
  \end{bmatrix}}
}_{\mathbf{D}(\mathbf{a},\mathbf{b}_3)}
\label{A_non_circular}
\end{align}
where $*$ denotes the element whose value we do not care about. We construct three CAUs $(h_1, h_2, h_3)$ with the following weight matrices,
\begin{align}
\underbrace{
  \scriptsize{
  \begin{bmatrix}
   0 & 0 & 0 & 0 & 0 \\
   1 & 0 & 0 & 0 & 0 \\
   0 & 1 & 0 & 0 & 0 \\
   0 & 0 & 1 & 0 & 0 \\        
   0 & 0 & 0 & 1 & 0 \\ 
  \end{bmatrix}}
}_{\mathbf{W}_1},
\underbrace{
  \scriptsize{
  \begin{bmatrix}
   1 & 0 & 0 & 0 & 0 \\
   0 & 1 & 0 & 0 & 0 \\
   0 & 0 & 1 & 0 & 0 \\        
   0 & 0 & 0 & 1 & 0 \\    
   0 & 0 & 0 & 0 & 1        
  \end{bmatrix}}
}_{\mathbf{W}_2},
\underbrace{
  \scriptsize{
  \begin{bmatrix}  
   0 & 1 & 0 & 0 & 0 \\
   0 & 0 & 1 & 0 & 0 \\        
   0 & 0 & 0 & 1 & 0 \\    
   0 & 0 & 0 & 0 & 1 \\
   0 & 0 & 0 & 0 & 0
  \end{bmatrix}}
}_{\mathbf{W}_3}
\label{A_non_circular}
\end{align}
respectively.
If $\mathbf{b}=\mathbf{b}_i$, then $h_i = 0$ and $h_{j\neq i}>0$ except for some special cases.  With WTA competition, $h_i'=1$ and $h_{j\neq i}'=0$. The translation variable $z$ can be inferred with simple decoding units $g(\mathbf{h}') = (-1,0,1) \cdot \mathbf{h}'= z$.

\subsection{Low-rank Approximation} \label{low-rank}

If $\mathbf{W}_k$ is large, we can approximate it in the following way.
For rank-one approximation, let
$\mathbf{W}_k = \mathbf{u}_{k}\mathbf{v}_{k}^T$, 
where $\mathbf{u}_{k}$ and $\mathbf{v}_{k}$ are a row of non-negative matrices $\mathbf{U}$ and $\mathbf{V}$, respectively. Then the right side of (\ref{cau}) can be written in the matrix form as
\begin{align} \mathbf{h}^* = 
\frac{1}{2}\Big[
(\mathbf{V}\mathbf{1}) \circ \mathbf{U}(\mathbf{a})^2 + (\mathbf{U}\mathbf{1}) \circ \mathbf{V}(\mathbf{b})^2\Big]  -  (\mathbf{U}\mathbf{a}) \circ (\mathbf{V} \mathbf{b}),
\label{rank_one}
\end{align}
where $\mathbf{1}$ is a vector of ones, $\circ$ is the element-wise multiplication and $(\cdot)^2$ is the element-wise square. The derivation can be found in Appendix. To obtain CAUs of higher ranks, we can apply sum-pooling over $\mathbf{h}^*$. That is, divide $\mathbf{h}^*$ into non-overlapping groups of equal size and sum the units in each group.

\section{Learning} \label{sec:learning}
To learn the non-negative weights in CAUs ($\mathbf{W}_k$ for full rank or $\mathbf{U}$ and $\mathbf{V}$ for rank-one), 
the conventional gradient descent based algorithms are not suitable because the nonnegativity of weights cannot be maintained after each update. Simply projecting weights onto the space of the nonnegative matrices after each update performs poorly in our experiments, where the learning converges at an extremely low speed or even diverges in many cases.

To address the above problem, we adopt a multiplicative update algorithm for the non-negative weight matrices in a neural network, which was originally used for non-negative matrix factorization \cite{lee2001algorithms}. 
For a non-negative matrix $\mathbf{W}$ in a neural network and loss function $E$, we decompose the gradient into two positive parts $\frac{\partial E}{\partial \mathbf{W}} = \nabla^+ - \nabla^-$,
where the two positive matrices $\nabla^+$ and $\nabla^-$ can be computed by 
\begin{align}
\nabla^+ &= \frac{1}{2}\left(\text{abs}\left(\frac{\partial E}{\partial \mathbf{W}}\right) + \frac{\partial E}{\partial \mathbf{W}}\right) + \epsilon, \\
\nabla^- &= \frac{1}{2}\left(\text{abs}\left(\frac{\partial E}{\partial \mathbf{W}}\right) - \frac{\partial E}{\partial \mathbf{W}}\right) + \epsilon,
\end{align}
where $\text{abs}(\cdot)$ is the element-wise absolute value and $\epsilon$ is a small positive scalar applied to each element of a matrix.
The multiplicative update algorithm is defined as 
\begin{align} 
\mathbf{W} &\leftarrow \mathbf{W}
 \circ 
\left(\frac{\nabla^- }{\nabla^+ }\right)^\eta,
\label{mul_update}
\end{align}
where $\eta$ is the learning rate hyperparameter and the division and the exponentiation are both element-wise. If $\mathbf{W}$ is initialized to be positive, the updated matrix will remain positive since all factors on the right hand side are positive. 

In practice, the multiplicative update algorithm can be used in a stochastic (or mini-batch) manner. Such use has been demonstrated in non-negative matrix factorization \cite{serizel2016mini}. Note that the multiplicative update still requires the gradients calculated by back-propagation. The gradient calculation for CAU can be found in Appendix.

\section{Related Work} \label{sec:related_work}
A classic model related to the proposed CAU is the 
energy model for motion detection \cite{adelson1985spatiotemporal} and stereo disparity \cite{fleet1996neural}. Each unit of the energy model computes the sum of squares of two Gabor filter outputs. No learning is involved in the energy model. There are models which compute the sum of squares of learnable filter outputs such as adaptive-subspace self-organized maps (ASSOM) \cite{kohonen1996emergence} and  independent subspace analysis (ISA) \cite{hyvarinen2000emergence}. Similar to CAU, ASSOM also has a competition mechanism (WTA).  However, the goal of both ASSOM and ISA is to learn appearance features which are invariant of image transformations, as different from our goal.
There is a line of research on relation learning based on Boltzmann machines \cite{memisevic2010learning,susskind2011modeling,huang2015conditional}. 
While Boltzmann machines allow a probabilistic formulation of relation inference, the training of Boltzmann machines is much more expensive compared to their non-probabilistic counterpart.
Non-negative weights have appeared in sum-product networks \cite{poon2011sum}, multi-layer perceptrons \cite{chorowski2015learning} and natural image statistics models \cite{hoyer2002multi,gutmann2013three}.
Our derivation of the low-rank approximation of CAUs follows from the low-rank approximation of bilinear units \cite{memisevic2010learning,kim2016hadamard}.

\section{Experiments} \label{sec:experiments}
We consider five fundamental image transformations for the relation learning tasks.
For image $\mathbf{x}$, its transformed image $\mathbf{y}$ is synthetically generated with ground-truth transformation parameters (or relation variables) $\mathbf{z}$. For each task, neural networks are trained in a supervised manner, given $\mathbf{x}$ and $\mathbf{y}$ as inputs and $\mathbf{z}$ as targets. We describe the details below.

\subsection{Tasks}
The five image transformations are: translation, rotation, scaling, affine transformation and projective transformation. They are called geometric transformations and can be unified as follows  \cite{hartley2003multiple}. An image can be transformed (or warped) by changing its coordinates. For each point of an image with homogeneous coordinates $\mathbf{p}=(p_1,p_2,1)$, the transformed point is $\mathbf{p}' = \mathbf{Hp}$ with homography matrix
\begin{align}
\mathbf{H}=
  \begin{bmatrix}
    h_{11} & h_{12} & h_{13} \\ 
    h_{21} & h_{22} & h_{23} \\ 
    h_{31} & h_{32} & h_{33} \\     
  \end{bmatrix}.
\end{align}
The type of the transformation depends on the parametrization of $\mathbf{H}$. Note that translation, rotation and scaling are special cases of affine transformation and affine transformation is a special case of projective transformation.  
We list the parametrization in each task and the range of the transformation parameters below.

\begin{itemize}
\vspace{0.5cm}
\item Translation. 
$\mathbf{z} \in [-5,5]^2$ and $\mathbf{H}=$
\begin{align*} 
\begin{bmatrix} 
1 & 0 & z_1 \\
0 & 1 & z_2 \\
0 & 0 & 1
\end{bmatrix}.
\end{align*}
Images are translated by $z_1$ pixels horizontally and $z_2$ pixels vertically. 

\vspace{0.5cm}
\item Rotation. $\mathbf{z} = z \in [-45,45]$, $\theta = \text{radian}(z)$ and $\mathbf{H}=$
\begin{align*}
\begin{bmatrix} 
\cos(\theta) & -\sin(\theta) & 0 \\
\sin(\theta) & \cos(\theta)  & 0 \\
0 & 0 & 1
\end{bmatrix}.
\end{align*}
Images are rotated by $z$ degrees. 

\vspace{0.5cm}
\item Scaling. 
$\mathbf{z} \in [0.5,2]^2$ and  $\mathbf{H}=$
\begin{align*}
\begin{bmatrix} 
z_1 & 0 & 0 \\
0 & z_2  & 0 \\
0 & 0 & 1
\end{bmatrix}.
\end{align*}
Images are scaled by $z_1$ horizontally and $z_2$ vertically. 

\vspace{0.5cm}
\item Affine. 
$\mathbf{z} \in [-0.5,0.5]^4$ and $\mathbf{H}=$
\begin{align*}
\begin{bmatrix} 
1 + z_1 & z_2      & 0 \\
z_3     & 1 + z_4  & 0 \\
0 & 0 & 1
\end{bmatrix}.
\end{align*}
To simplify the problem, we discard translation such that $h_{13}$ and $h_{23}$ are zero.

\vspace{0.5cm}
\item Projective. 
$\mathbf{z} \in [-0.5,0.5]^4\times[-0.01,0.01]^2$
and $\mathbf{H}=$
\begin{align*}
\begin{bmatrix} 
1 + z_1 & z_2      & 0 \\
z_3     & 1 + z_4  & 0 \\
z_5 & z_6 & 1
\end{bmatrix}.
\end{align*}
$z_5$ and $z_6$ are much more sensitive than the other variables. Hence, their range is set to be smaller. Larger range will cause extremely distorted images. To simplify the problem, we discard translation such that $h_{13}$ and $h_{23}$ are zero.
\end{itemize}

\subsection{Data}
We generate training image patches from the gray-scaled
CIFAR-10 dataset\footnote{\url{https://www.cs.toronto.edu/~kriz/cifar.html}} \cite{krizhevsky2009learning}. 
For each task and for each image in CIFAR-10, we apply an image transformation with $\mathbf{z}$ randomly sampled from uniform distributions to obtain an image pair. Then we crop an image patch of size 11$\times$11 at the center of each image of the image pair. Repeat the process 10 times. With this procedure, we obtain a training set of size 500,000 and a testing set of size 100,000 for the relation learning tasks.

\subsection{Models}
We test three neural network models, each of which uses a different type of relation units: concatenation, bilinear and CAU. For all the models, $f(\cdot)$ is the identity function and $g(\cdot)$ is a multi-layer perceptron. We call the three neural network models,  concatenation network (CTN), bilinear network (BLN) and \textit{contrast association network} (CAN), respectively. 
For BLN and CAN, we use low-rank approximation of the bilinear units and CAUs, as described in Section \ref{low-rank}.
In BLN, we apply $l_2$ normalization, which empirically performs better than softmax (or softmin), on the outputs of bilinear units. We also experimented with the non-negative constraint on the bilinear units but found it performs poorly and therefore discarded it.
For fairness of the comparison, all three models are set to have essentially the same size in each task. To test the generality, all models are set to have the same size in different tasks, except that the last layer depends on the dimensionality of the targets $\mathbf{z}$ in the task. The models are specified in Table \ref{models}.

\begin{figure}[t]
\centering
\subfloat[Translation]{
\includegraphics[width=0.8\columnwidth]{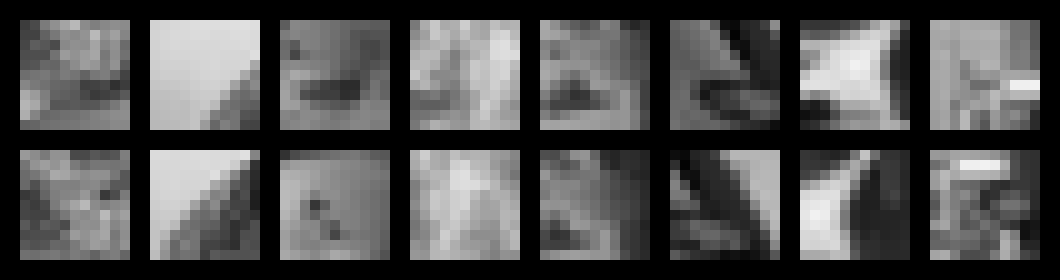}}

\subfloat[Rotation]{
\includegraphics[width=0.8\columnwidth]{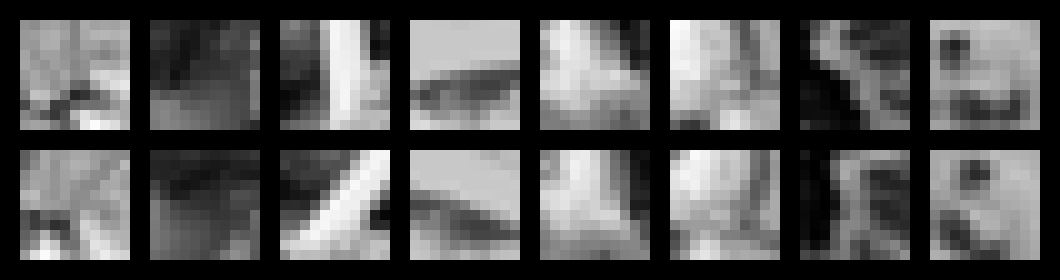}}

\subfloat[Scaling]{
\includegraphics[width=0.8\columnwidth]{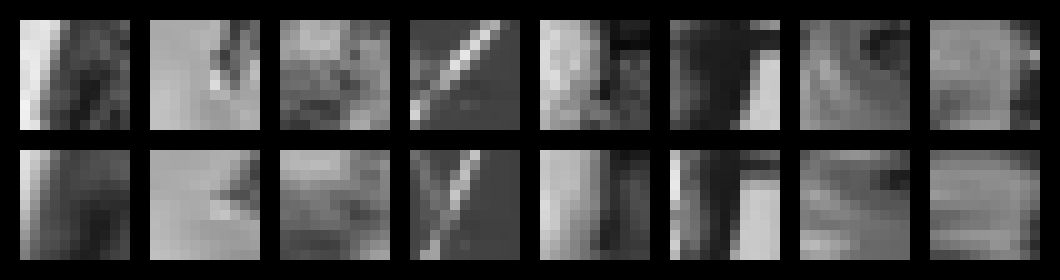}}

\subfloat[Affine]{
\includegraphics[width=0.8\columnwidth]{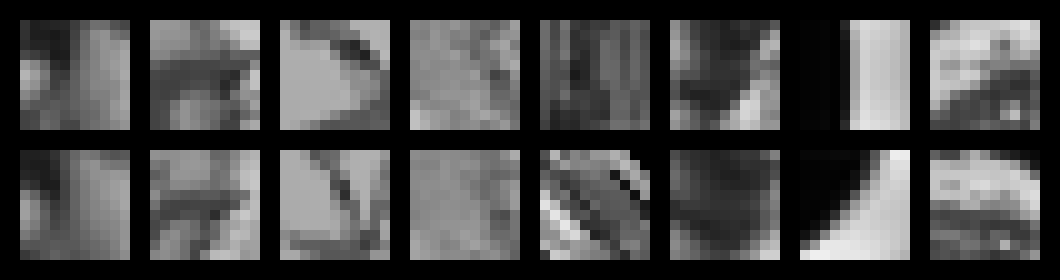}}

\subfloat[Projective]{
\includegraphics[width=0.8\columnwidth]{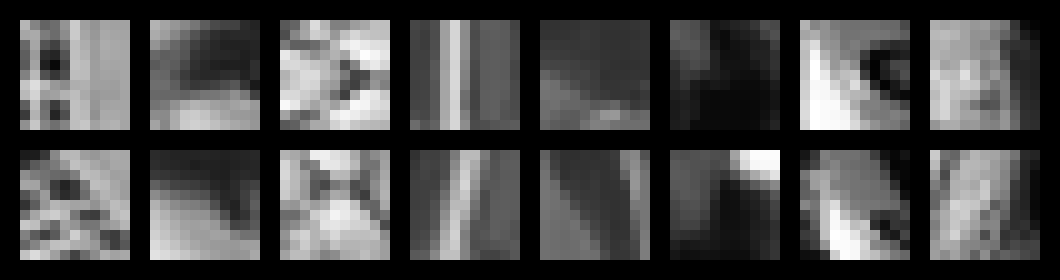}}
\caption{Sample image patches. For (a)-(e), the first row contains the original image patches $\mathbf{x}$ and the second row contains the corresponding transformed image patches $\mathbf{y}$.}
\end{figure}

\begin{table}[h!]
\caption{Model specification. The size of each layer is on the right, if applicable. Concat denotes concatenation. Linear denotes fully connected layer. PReLU denotes the parametric ReLU activation function \cite{he2015delving}. * denotes rank-one approximation. The sum-pooling pools every 4 elements of a vector. dim($\mathbf{z}$) denotes the dimensionality of $\mathbf{z}$.}
\label{models}
\begin{center}
\begin{small}
\begin{tabular}{ccc}
\hline
CTN & BLN & CAN \\
\hline
Concat      & Bilinear*, 1200       & CAU*, 1200    \\
Linear, 1200 & Sum-Pooling, 300      & Sum-Pooling, 300      \\
PReLU       & $l_2$ Norm         & Softmin\\
Linear, 300  & Linear, 100&  Linear, 100 \\
PReLU        & PReLU            & PReLU \\
Linear, 100  &  Linear, 100         & Linear, 100 \\ 
PReLU        & PReLU           & PReLU  \\
Linear, 100  & Linear, dim($\mathbf{z}$) & Linear, dim($\mathbf{z}$)\\ 
PReLU       \\
Linear, dim($\mathbf{z}$)   \\
\hline
\end{tabular}
\end{small}
\end{center}
\end{table}

\subsection{Settings}

All image patches are of size 11$\times$11 and are reshaped to vectors of size 121. The numerical range of each element of the vectors is normalized from $[0, 255]$ to $[-0.5, 0.5]$ by dividing 255 and then subtracting 0.5. This normalization significantly accelerates the training.

We use the mean-squared-error (MSE) as the loss function.

All the parameters of all models are initialized randomly according to uniform distributions.
We use the multiplicative update algorithm (described in Section \ref{sec:learning}) for non-negative weights ($\mathbf{U}$ and $\mathbf{V}$) and Adam \cite{kingma2014adam} for unconstrained weights. 

For all models in all tasks, we use the same training hyperparameter setting. The initial learning rates are $\eta=0.005$ (multiplicative update) and  $\alpha=0.005$ (Adam). Both are multiplied by 0.95 for every 500 mini-batch updates. The size of each mini-batch is 100. There are 200,000 mini-batch updates in total. $\epsilon = 10^{-20}$ in the multiplicative update. 

We use MATLAB\footnote{\url{https://www.mathworks.com/}} for generating the data and Torch\footnote{\url{http://torch.ch/}} for neural network training and testing.

\subsection{Results}
We use two measures of error to evaluate our results.

\vspace{0.1in}
\begin{itemize}
\item Parameter error. It is defined as the MSE between the ground-truth transformation parameters $\mathbf{z}$ and the inferred parameters $\widehat{\mathbf{z}}$, that is,  $\| \mathbf{z} - \widehat{\mathbf{z}}\|_2$. An advantage of parameter error is interpretability. For example, in inferring the rotation between two images, it is desirable that the inference error is measured in degrees. A disadvantage is that it is difficult to compare the inference error across tasks since different tasks have different range of transformation parameters.

\vspace{0.1in}
\item Transformation error. Define four points with homogeneous coordinates
\begin{align*}
\mathbf{p}_1 = (0,0,1),\  \mathbf{p}_2 &= (1,0,1), \\ 
\mathbf{p}_3 = (1,1,1),\  \mathbf{p}_4 &= (0,1,1).
\end{align*}
Let 
$\mathbf{p}_i' = \mathbf{Hp}_i$ be the transformed point with the ground-truth homography $\mathbf{H}$ and let $\widehat{\mathbf{p}}_i' = \widehat{\mathbf{H}}\mathbf{p}_i$ be the transformed point with the inferred homography $\widehat{\mathbf{H}}$. The transformation error is defined as 
\begin{align}
\frac{\sum_{i=1}^4 \|\mathbf{p}_i' - \widehat{\mathbf{p}}_i'\|_2}{\sum_{j=1}^4\|\mathbf{p}_j'\|_2}.
\end{align}
It is scale-invariant in the sense that the error is unchanged if we multiply $\mathbf{H}$ and $\widehat{\mathbf{H}}$ by a non-zero constant. Therefore it is more suitable to compare this inference error across different tasks.
\end{itemize}
\vspace{0.1in}

The results are listed in Tables \ref{VE} and \ref{TE}, with errors averaged over the testing sets. We can see that CAN achieves the lowest errors in every task. From Table \ref{TE}, we can see that the transformation error of CAN goes up when the complexity of the tasks is increased.

\begin{table}[h!]

\begin{center}
\caption{Mean parameter error on testing sets.}
\label{VE}
\begin{small}
\begin{tabular}{lcccc}
\hline
Task &   CTN & BLN & CAN \\
\hline
Translation &   0.773 & 1.893 & \textbf{0.049} \\
Rotation &   9.854 & 5.925 & \textbf{3.518}   \\
Scaling &  0.018 & 0.025 & \textbf{0.017}     \\
Affine &   0.014 & 0.020 & \textbf{0.010}  \\
Projective &  \textbf{0.030}  & 0.032 & \textbf{0.030} \\
\hline
\end{tabular}
\end{small}
\end{center}

\begin{center}
\caption{Mean transformation error on testing sets.}
\label{TE}
\begin{small}
\begin{tabular}{lcccc}
\hline
Task  & CTN & BLN & CAN \\
\hline
Translation &   0.198 & 0.322 & \textbf{0.014} \\
Rotation &   0.025 & 0.018 & \textbf{0.014}   \\ 
Scaling &  0.056 & 0.065 & \textbf{0.055}     \\
Affine &  0.084 & 0.100 & \textbf{0.072}  \\
Projective & 0.160 & 0.166 & \textbf{0.158} \\
\hline
\end{tabular}
\end{small}
\end{center}

\end{table}

\section{Dicussion}\label{sec:conclusion}
In our present work, the experiments are limited to small image patches and to simple image transformations. It is not clear how well CAUs perform on whole images and more complex tasks. Future work includes extending CAUs to handle whole images and more complex tasks such as three-dimensional transformations.

\section*{Appendix}

\subsection*{Rank-One Approximation of CAU}

Since $\mathbf{W}_k = \mathbf{u}_{k}\mathbf{v}_{k}^T$, we have
\begin{align}
h_k &= \frac{1}{2}\sum_{ij}W_{ijk}(a_i-b_j)^2  \\
&= \frac{1}{2}\sum_{ij}W_{ijk}(a_i^2+b_j^2) -\sum_{ij}W_{ijk}a_ib_j \\
&= \frac{1}{2}\Big[\mathbf{1}^T\mathbf{W}^T_k(\mathbf{a})^2  +  \mathbf{1}^T\mathbf{W}_k(\mathbf{b})^2\Big] - \mathbf{a}^T \mathbf{W}_k \mathbf{b}  \\
&= \frac{1}{2}\Big[\mathbf{1}^T\mathbf{v}_k\mathbf{u}^T_k(\mathbf{a})^2  +  \mathbf{1}^T\mathbf{u}_k\mathbf{v}^T_k(\mathbf{b})^2\Big] - \mathbf{a}^T \mathbf{u}_k\mathbf{v}^T_k \mathbf{b}.
\end{align}
In matrix form
\begin{align}
\mathbf{h} &= \frac{1}{2}\Big[
(\mathbf{V}\mathbf{1}) \circ \mathbf{U}\mathbf{a}^2 + (\mathbf{U}\mathbf{1}) \circ \mathbf{V}\mathbf{b}^2\Big] - (\mathbf{U}\mathbf{a}) \circ (\mathbf{V} \mathbf{b}).
\end{align}

\subsection*{Gradients of CAU}

\begin{align}
\frac{\partial E}{\partial \mathbf{a}} 
&= \sum_k\frac{\partial E}{\partial h_k}\frac{\partial h_k}{\partial \mathbf{a}}, \quad
\frac{\partial E}{\partial \mathbf{b}} 
= \sum_k\frac{\partial E}{\partial h_k}\frac{\partial h_k}{\partial \mathbf{b}}, \\
\frac{\partial h_k}{\partial \mathbf{a}}
&= (\mathbf{W}_k^T\mathbf{1})\circ\mathbf{a}  -  \mathbf{W}_k \mathbf{b}, \\
\frac{\partial h_k}{\partial \mathbf{b}}
&= (\mathbf{W}_k\mathbf{1})\circ\mathbf{b}  -  \mathbf{W}_k^T \mathbf{a},  \\
\frac{\partial E}{\partial \mathbf{W}_k} 
&= \frac{\partial E}{\partial h_k}\frac{\partial h_k}{\partial \mathbf{W}_k}, \\
\frac{\partial h_k}{\partial \mathbf{W}_k}
&= \frac{1}{2}\Big[(\mathbf{a}^2)\mathbf{1}^T + \mathbf{1}(\mathbf{b}^2)^T\Big]  - \mathbf{a} \mathbf{b}^T.  
\end{align}
For $\mathbf{W}_k = \mathbf{u}_k\mathbf{v}_k^T,$
\begin{align}
\frac{\partial E}{\partial \mathbf{u}_k} 
&= \frac{\partial E}{\partial h_k}\frac{\partial h_k}{\partial \mathbf{u}_k}, \quad \frac{\partial E}{\partial \mathbf{v}_k} 
= \frac{\partial E}{\partial h_k}\frac{\partial h_k}{\partial \mathbf{v}_k}, \\
\frac{\partial h_k}{\partial \mathbf{u}_k}
&= \frac{1}{2}\Big[(\mathbf{v}_k^T\mathbf{1})\mathbf{a}^2  +  (\mathbf{v}^T_k\mathbf{b}^2)\mathbf{1}\Big] - (\mathbf{v}^T_k \mathbf{b})\mathbf{a}, \\
\frac{\partial h_k}{\partial \mathbf{v}_k}
&= \frac{1}{2}\Big[(\mathbf{u}_k^T\mathbf{1})\mathbf{b}^2 + (\mathbf{u}^T_k\mathbf{a}^2)\mathbf{1} \Big] - ( \mathbf{u}_k^T\mathbf{a})\mathbf{b}.
\end{align}

\bibliography{CAN_IJCNN}
\bibliographystyle{abbrv}

\end{document}